\documentclass[runningheads]{llncs}
\usepackage{graphicx}
\usepackage{amsmath,amssymb,amsfonts}
\usepackage{algorithmic}
\usepackage{graphicx}
\usepackage{textcomp}
\usepackage{xcolor}
\usepackage{algorithm}
\usepackage[natbib = true, doi = false, isbn = false]{biblatex}
\usepackage{pifont}
\usepackage{array}
\usepackage{ifthen}
\usepackage{subfigure}
\def\BibTeX{{\rm B\kern-.05em{\sc i\kern-.025em b}\kern-.08em
    T\kern-.1667em\lower.7ex\hbox{E}\kern-.125emX}}
    
\def\transpose{\perp}    
\def\learningrate{\alpha}
\def\weight{w}
\def\xstar{x_\star}
\def\ystar{y_\star}
\def\weightstar{w_\star}

\def\refb#1{(\ref{#1})}

\DeclareMathOperator{\Mat}{Mat}

\def\mN{{\mathbb{N}}}

\newif\ifextended
\extendedtrue

\newcolumntype{C}[1]{>{\centering\arraybackslash}p{#1}}
\newboolean{extendedversion}
\setboolean{extendedversion}{false}
    
\newcommand{\norm}[1]{\left\lVert#1\right\rVert}
\newcommand{\N}{\mathbb{N}} 
\newcommand{\R}{\mathbb{R}} 

\begin{document}
\title{Non-Convergence and Limit Cycles in the Adam optimizer\thanks{This paper presents results of the project "LeaP – Learning Poses" supported by the Bavarian Ministry of Science and Art under Kap. 15 49 TG 78.}}
%
%
\author{Sebastian Bock\inst{1}\and
Martin Wei{\ss}\inst{1}}
\authorrunning{S. Bock and M. Wei{\ss}}
%
\institute{OTH Regensburg, Prüfeninger Str. 58, 93049 Regensburg, Germany \email{\{sebastian2.bock,martin.weiss\}@oth-regensburg.de}}
\maketitle              
\begin{abstract}
One of the most popular training algorithms for deep neural networks is the Adaptive Moment Estimation (Adam) introduced by Kingma and Ba. 
Despite its success in many applications there is no satisfactory 
convergence analysis: only local convergence can be shown for batch mode under some
restrictions on the hyperparameters, counterexamples
exist for incremental mode. Recent results show that for simple
quadratic objective functions limit cycles of period 2 exist in batch mode, 
but only for atypical hyperparameters, and only for the algorithm without bias correction. 
We extend the convergence analysis for Adam in the batch mode with bias correction
and show that
even for quadratic objective functions as the simplest case of convex functions
2-limit-cycles exist, for all choices of the hyperparameters. 
We analyze the stability of these limit cycles and relate our analysis to other
results where approximate convergence was shown, but under the additional assumption of
bounded gradients which does not apply to quadratic functions.
The investigation heavily relies on the use of computer algebra
due to the complexity of the equations.
\end{abstract}

\section{Introduction}

Adaptive Moment Estimation (Adam), originally presented by Kingma and Ba 
\cite{Kingma.2014} is probably the most widely used training algorithm for neural
networks, especially convolutional neural networks. Implementations exist in all
popular machine learning frameworks like Tensorflow or PyTorch. Despite its apparent success the
theoretical basis is weak: 
The original proof in \cite{Kingma.2014} is wrong as has been noted by several authors,
see \cite{Bock.2017, Rubio.2017}. Of course a faulty proof does not imply that
the Adam optimizer does not converge, and indeed local convergence can be shown
for batch mode 
under reasonable restrictions on the hyperparameters, see \cite{BockWeiss.2019}.
Furthermore \cite{Reddi.2018} give an example in incremental mode where the
regret does not converge, neither do the arguments of the objective function.

Several results exist which show $\varepsilon$-bounds on the gradients, 
$\norm {\nabla f(\weight_t)} < \varepsilon$ for $\varepsilon>0$ arbitrarily small for all
$t$ is sufficiently large, see \cite[Theorem 3.4.]{De.2018}, \cite{Chen.2018},
\cite[Theorem 3.3]{Zhou.2018}. 
Other results show asymptotic bounds on the regret or 
that the function values come close to the minimum, $f(\weight_t)-f(\weightstar) < \varepsilon$, see \cite{Chen.2018} for example. 
To the best of the authors' knowledge, \cite{BockWeiss.2019} is the only 
(partial) result on weight convergence in the standard mathematical definition
$\lim_{t\to\infty} \weight_t =\weightstar$, however only in a local sense.

Contrary to these results \cite{daSilva.2018}  shows that 2-cycles exist for
the Adam optimizer without bias correction for the simple case of a scalar 
quadratic objective function
$f(\weight)=\frac 1 2 \weight^2$. Quadratic objective functions are a natural 
benchmark for any optimization algorithm in convex analysis: The standard gradient
descent algorithm converges for learning rate small enough, see \cite{Nesterov.2004},
so this behaviour should be replicated by more sophisticated gradient motivated 
adaptive algorithms like Adam.
However \cite[Proposition 3.3]{daSilva.2018} only deals with the case $\beta_1=0$ which means that the first moments are not adapted at all -- this case hardly can be called
Adam any more. 

We extend the results of \cite{daSilva.2018} to the general case of
hyperparameters $\alpha>0$, $0<\beta_1< 1$, $0<\beta_2< 1$, and show existence
of 2-limit-cycles for scalar objective functions $f(\weight)=\frac 1 2 c\weight^2$, 
$c>0$, which easily generalizes by diagonalization to strictly convex quadratic functions 
$f(\weight)=\frac 1 2 \weight^\transpose C \weight$, $C$ positive definite.
This is done for the Adam algorithm in batch mode only, but also for bias correction.

We give numerical evidence that for typical values of $\beta_1, \beta_2$ near 1
these 2-cycles are unstable, and stable for $\beta_1, \beta_2$ near 0.
The analysis of the limit cycles is not exhaustive: more 2-cycles may exist, 
and cycles of larger period. However our results suffice to clarify the global 
non-convergence of Adam even for strictly convex functions under the fairly 
standard assumptions of bounds on the Hessian like 
$0 < l I_n \leq \nabla^2 f(\weight) \leq LI_n$ for all $\weight\in \R^n$.

The outline of the paper is as follows: In Section 2 we define our variant of
the Adam algorithm and explain the steps of our proof. In Section 3 these steps
are carried out. The Maple code used can be obtained from the publishers web site\footnote{- URL missing - We can provide the code for the referees of course}.
Section 4 shows numerical simulations suggesting that a Hopf
bifurcation occurs, before we state some conclusions and relate our results to
other research. 

{\em Notation:}
With $\Mat_n$ we denote the set of all real $n$-by-$n$ matrices. 
The symbol $\transpose$ denotes the transpose of a vector or matrix. 
The class of $k$-times continuously differentiable functions from $\R^n$ to $\R^m$ is denoted by $C^k(\R^n,\R^m)$, with $\nabla f$ the gradient and $\nabla^2 f$ the Hessian
for scalar valued functions. 
Throughout this paper we assume $f:\R^n\to\R^n$ at least $C^1$, $C^2$ for some results. 
In the numerical tests we denote the machine accuracy with $eps = 2.2204*10^{-16}$, i.e. standard IEEE floating point numbers with double 
precision.

\section{Outline of the Proof}
\subsection{Definition of the Algorithm}

In the course of time, variations of the Adam optimizer were developed.
We use the version of Adam shown in Algorithm \ref{alg:ADAM} (The symbols $\otimes, \oplus$ and $\oslash$ denote the component-wise multiplication and division of vectors, as well as component-wise addition of vectors and scalar.). The main points worth noting are:

Originally Kingma and Ba \cite{Kingma.2014} use $\sqrt{v} \oplus \varepsilon$ and the bias correction in $\hat{m}$ and $\hat{v}$. Other publications like \cite{Reddi.2018, Chen.2018} do not use an $\varepsilon$ to avoid division by zero 
as well as \cite{Zou.2018}, but the latter initialize $v_0 = \varepsilon$ with essentially the same effect. We use the variant with $\varepsilon$ in the denominator; otherwise the initial value $v_0=0$ would have to be excluded in all results, and one could
not talk about stability of a fixed point $\weight_\star$ of the iteration 
corresponding to a minimum of the objective function.

Also we use $\sqrt{v \oplus \varepsilon}$ as in \cite{daSilva.2018, BockWeiss.2019},
instead of $\sqrt{v}\oplus \varepsilon$ as in the original publication \cite{Kingma.2014}. 
Our variant has the advantage that the iteration is continuously differentiable
for all $v\geq0$ whereas the $\varepsilon$ outside of the square root leads to a non differentiable exception set. 
The numerical differences of the two variants are marginal and described in more detail in \cite{BockWeiss.2019}.
 
All cited publications with the exception of \cite{daSilva.2018} 
apply a bias correction in the learning rate $\learningrate_t$; we use the same 
bias correction as described in \cite[Section 2]{Kingma.2014}.

\begin{algorithm}
\caption{Adam Optimization}
\label{alg:ADAM}
\begin{algorithmic}[1]
\REQUIRE $\learningrate \in \R^+$, $\varepsilon \in \R$ $\beta_1, \beta_2 \in (0,1)$, $\weight_0 \in \R^n$ and the function $f(\weight) \in C^2 \left( \R^n, \R \right)$
\STATE $m_0 = 0$
\STATE $v_0 = 0$
\STATE $t = 0$
\WHILE{$\weight$ not converged}
\STATE $m_{t+1} = \beta_1 m_t + (1-\beta_1) \nabla_w f(\weight_t) $ 
\STATE $v_{t+1} = \beta_2 v_t + (1- \beta_2) \nabla_w f(\weight_t) \otimes \nabla_w f(\weight_t)$
\STATE $\weight_{t+1} = \weight_t - \learningrate \frac{\sqrt{1-\beta_2^{t+1}}}{\left( 1-\beta_1^{t+1}\right)} m_{t+1} \oslash \sqrt{v_{t+1} \oplus \varepsilon}$
\STATE $t = t+1$
\ENDWHILE
\end{algorithmic}
\end{algorithm}

We denote $x=(m,v,\weight)$ the state of the Adam iteration, and interpret the
algorithm as discrete time dynamical system $x_{t+1}=T(t, x_t; p)$ with
$T = T(t, x; p) =   T(t, m,v,w; \alpha, \beta_1, \beta_2, \varepsilon)$ 
to express the dependence of the iteration on the state 
$x = (m,v,\weight)$ and the 
hyperparameters $p=(\alpha, \beta_1, \beta_2, \varepsilon)$. 

We write Adam without bias correction in the same way as
$\bar T = \bar T(x; p)  
= \bar T(m,v,w; \alpha, \beta_1, \beta_2, \varepsilon)$. This gives an autonomous 
dynamical system; the right hand side does not explicitly depend on $t$. The difference
between the two systems is denoted by $\Theta(t,x;p)$, so we have analogous to 
\cite{BockWeiss.2019}
\begin{align}
\label{eq:System_Origin}
x_{t+1} = T(t,x_t;p ) = \bar{T} ( x_t ;p) + \Theta ( t, x_t,p )
\end{align}
with
\begin{align}
\label{eq:AutonomousSystemAdam}
\bar{T} \left( x_t ;p \right) &= \begin{bmatrix}
\beta_1 m_t + \left( 1-\beta_1 \right)g \left( \weight_t \right)\\
\beta_2 v_t + \left( 1-\beta_2 \right)g \left( \weight_t \right) \otimes g \left( \weight_t \right) \\
\weight_t - \learningrate \left(m_{t+1} \oslash \sqrt{v_{t+1} \oplus \varepsilon}\right)
\end{bmatrix}
\end{align}
and
\begin{align}
\label{eq:NonAutonomousSystemAdam}
\Theta ( t,x_t;p) &= 
\begin{bmatrix}
0\\
0\\
-\learningrate \left(\frac{\sqrt{1-\beta_2^{t+1}}}{1-\beta_1^{t+1}} -1 \right) \left(m_{t+1} \oslash \sqrt{v_{t+1} \oplus \varepsilon}\right)
\end{bmatrix}
\end{align}

\subsection{Steps of the Proof}

We show that for the objective  function $f(\weight) = \frac 1 2 c \weight^2$ with 
$c>0$ 
2-cycles occur for a wide range of hyperparameters in the Adam iteration 
without bias correction, and that iterations of the bias corrected algorithm
converge to this limit cycle if it is stable. We proceed
in several steps, analyzing simplified variants of Adam first, then adding complexity
in each step. The analysis uses Maple as a computer algebra system and some continuity
 and disturbance arguments because the naive approach of applying the {\tt solve} 
command to find 2-cycles fails -- the equations  are too complicated. 

\begin{enumerate}
\item
We start with the scalar case $f(\weight) = \frac 1 2 c \weight^2$, $c>0$.
We obtain analytical expressions for 2-limit-cycles 
$\bar T^2(x; \alpha, \beta_1, \beta_2, \varepsilon) = x$ 
of the autonomous system with $\varepsilon=0$. 
\item
Calculating the eigenvalues of $\bar T^2$ we find that these do not depend on the 
learning rate $\alpha$ and the factor $c$ in the objective function.
For some typical values of the 
hyperparameters we give evidence that these limit cycles are often attractive. 
We have not managed to give analytical estimates for stable eigenvalues using CAS so far.
\item
Using the implicit function theorem we show that for a neighbourhood
of $\varepsilon_\star=0$ there exists a unique limit cycle of the autonomous system with $\varepsilon>0$.
By continuity of the eigenvalues, these limit cycles are also attractive for $\varepsilon$ small enough. 
\item 
We apply a disturbance estimate to show that locally solutions of $T(t, x; p)$  converge to
the limit cycles of $\bar T(x; p)$. The proof is essentially the same as in \cite[Theorem V.1.]{BockWeiss.2019} and holds for cycles of any integer length.
\end{enumerate}

\section{Existence of 2-Limit-Cycles in Adam}

Step 1: We show that limit cycles of period 2 exist for Adam without bias correction, 
i.e. the autonomous system $\bar T$. A 2-cycle corresponds to a non-constant solution of
$\bar T(\bar T(x;p);p))=x$, so we try to solve this system of equations with Maple. This 
fails, so we do not use arbitrary parameters but fix $\varepsilon=0$. Now Maple succeeds
and returns
\begin{equation}
\label{eq:wtildeExplizit}
  \tilde m=\frac 1 2 {\frac {c \left( {\it \beta_1}-1 \right) ^{2}\alpha}{ \left( {
\it \beta_1}+1 \right) ^{2}}},
\tilde v=\frac 1 4\,{\frac {{\alpha}^{2} \left( \beta_1^{2}-
2\,{\it \beta_1}+1 \right) {c}^{2}}{ \left( {\it \beta_1}+1 \right) ^{2}}},
\tilde \weight=\frac 1 2
\,{\frac {\alpha \left( {\it \beta_1}-1 \right) }{{\it \beta_1}+1}} 
\end{equation}
with $(-\tilde m, \tilde v, -\tilde \weight)$ the other point on the 2-cycle. 
Note that $\tilde m\neq 0$, so we have a 2-cycle indeed. 
We abbreviate these points as $\tilde x_1$ and $\tilde x_2$.
The $v$ components of $\tilde x_1$ and $\tilde x_2$ are identical.
This limit 2-cycle exists for all $\beta_1\neq \pm 1$, that is for all reasonable Adam hyperparameters. Maple also returns more 2-cycles depending on the
roots of $2\beta_1\beta_2-2\beta_2^2-2\beta_1+2\beta_2$, we have not analyzed these. 
We could not determine cycles of greater period $q\in \mN$ by solving $\bar T^q(x)=x$.

Step 2:
However we can investigate the stability of the limit cycle. This is done using
the Eigenvalues of the Jacobian of $\bar T^2(\tilde x_1)$, which are the same 
as those of $\bar T^2(\tilde x_2)$.
The Jacobian is computed by Maple as
\[
 \left[ \begin {array}{ccc} 
  - \left( {\it \beta_1}+2 \right) {\it \beta_1}&2{
\frac { \left( {\it \beta_1}+1 \right) {\it \beta_2}}{{\it \alpha}{\it c}}}&{\it 
c} \left( {\it \beta_1}-1 \right)  \left( {\it \beta_1}+2{\it \beta_2}-1
 \right) \\ \noalign{\medskip}-2 \left( {\it \beta_2}-1 \right) {\it \alpha}
{\it \beta_1}{\it c}&3{{\it \beta_2}}^{2}-2{\it \beta_2}&2{\frac {{{\it c
}}^{2} \left( {\it \beta_1}-1 \right)  \left( {\it \beta_1}+3/2{\it \beta_2}-1/2
 \right)  \left( {\it \beta_2}-1 \right) {\it \alpha}}{{\it \beta_1}+1}}
\\ \noalign{\medskip}2{\frac {{\it \beta_1} \left( {\it \beta_1}+1 \right) 
 \left( 1 - {\it \beta_1}-2{\it \beta_2} \right) }{{\it c} \left( {\it \beta_1}-1
 \right) }}&2{\frac { \left( 2{\it \beta_1}+3{\it \beta_2}-1 \right) 
 \left( {\it \beta_1}+1 \right) {\it \beta_2}}{{{\it c}}^{2}{\it \alpha} \left( {
\it \beta_1}-1 \right) }}&2{{\it \beta_1}}^{2}+ \left( 8{\it \beta_2}-6 \right) {
\it \beta_1}+6 {{\it \beta_2}}^{2}-6{\it \beta_2}+1
\end {array} \right] 
\]
and not easy to interpret.  
Using Maple we obtain a very lengthy expression for the eigenvalues which does 
{\em not} depend on $\alpha$ or $c$. Details can be seen in the supplementary code.
This is surprising as most algorithms show dependence on the learning rate,
and one might assume that the behaviour at a limit cycle is different at least for 
$\alpha\to 0$ and $\alpha\to \infty$.
So the stability of the limit cycle depends on $\beta_1$ and $\beta_2$ only.
If the limit cycle is unstable there are still chances that the Adam algorithm
converges. 

\begin{figure}[!t]
\begin{subfigure}{}
  \includegraphics[trim = 50 270 50 270, clip, width=0.5\textwidth]{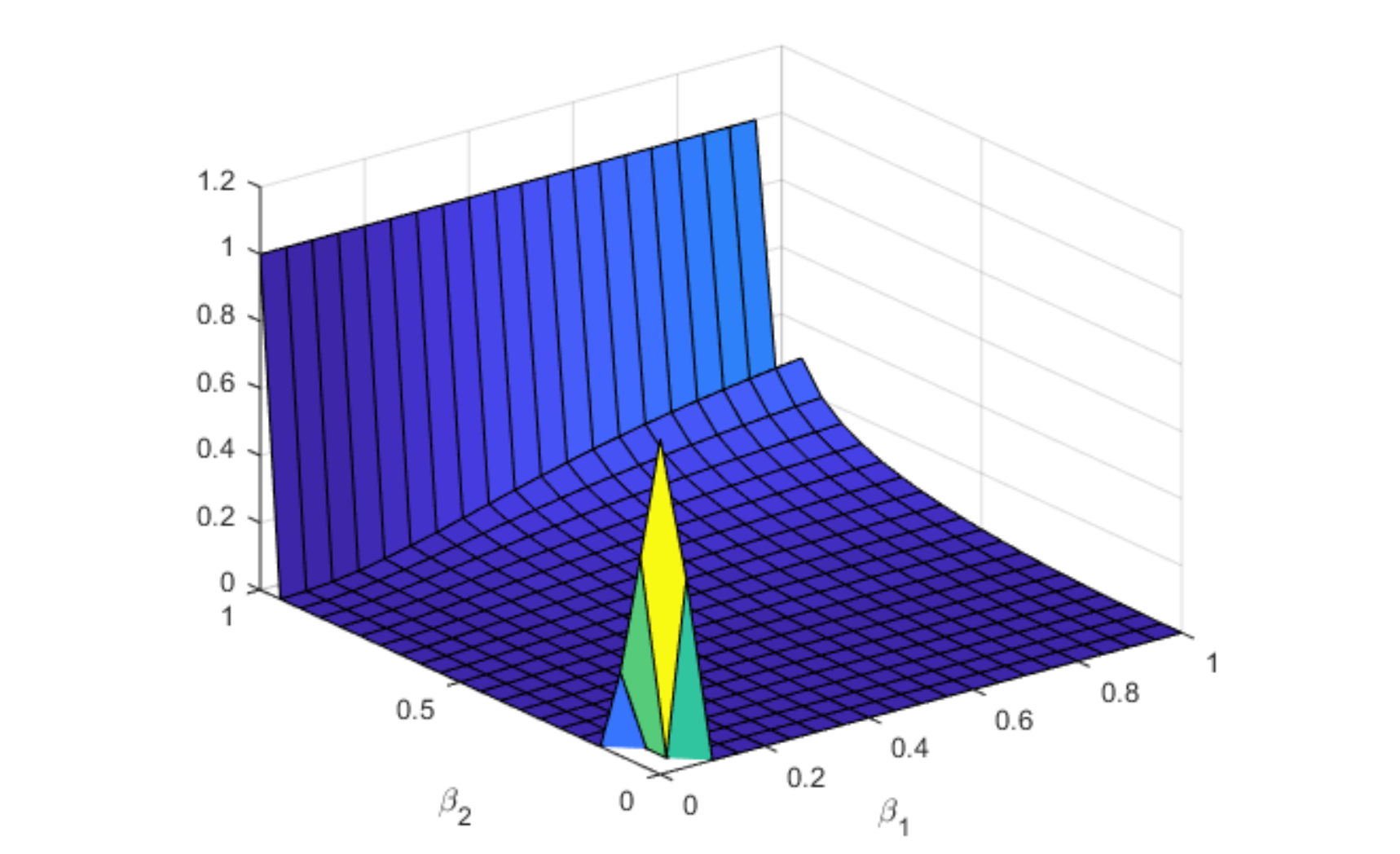}
  \end{subfigure}
  \begin{subfigure}{}
  \includegraphics[trim = 50 270 50 270, clip, width=0.5\textwidth]{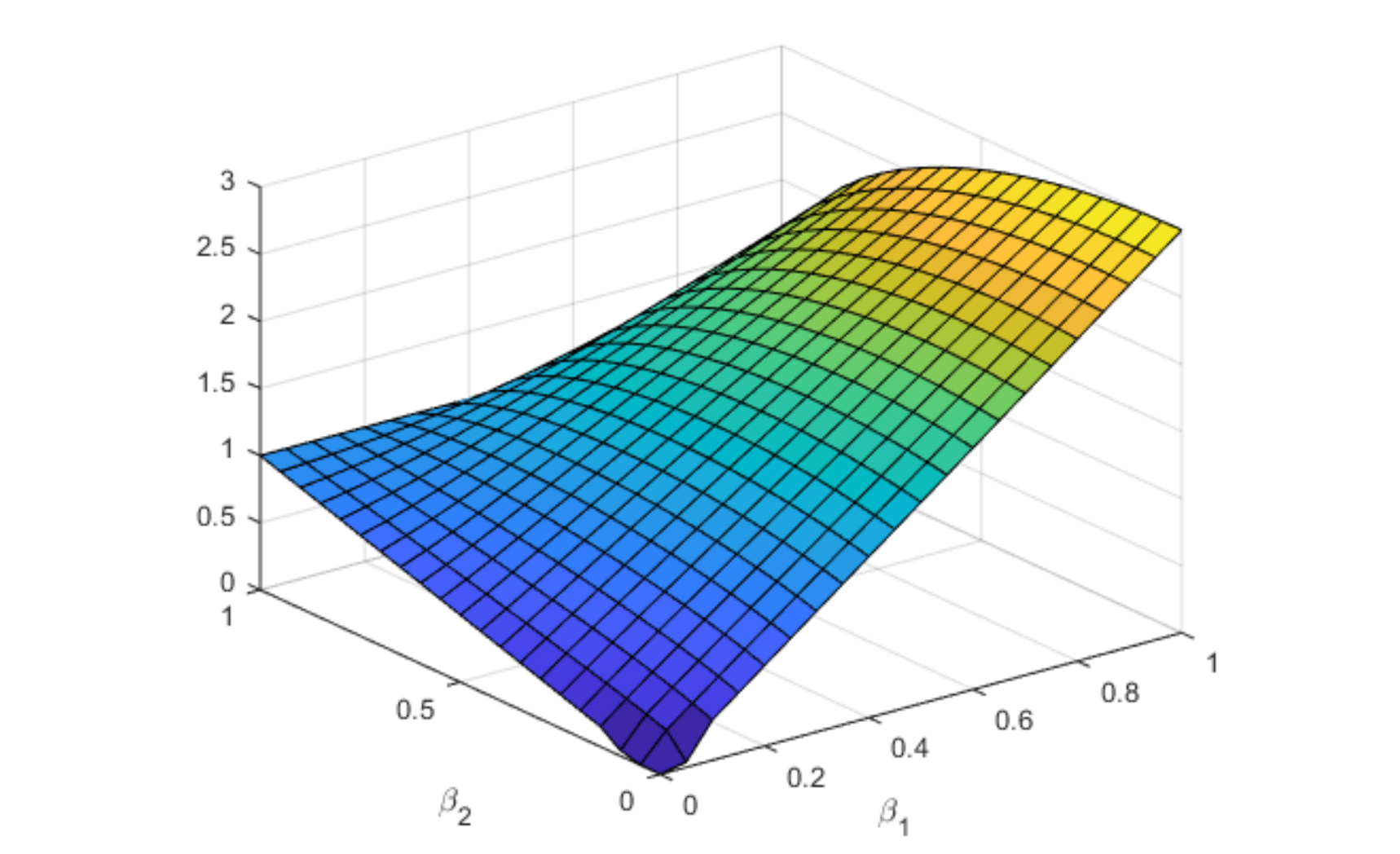}
  \end{subfigure}
\caption{Absolute magnitude of real and complex eigenvalues}
\label{fig:AbsEigenwerte}
\end{figure}

Plotting the absolute values of the eigenvalues over $\beta_1$ and $\beta_2$ we can see
that these limit cycles are often attractive, see Figure \ref{fig:AbsEigenwerte}. 
Local attractivity holds if the absolute values are less than 1: This is
the case for the real eigenvalue, see left plot, but the magnitude approaches 1 as  
$\beta_2$ approaches 1 -- consider the standard value $\beta_2=0$ suggested in 
\cite{Kingma.2014}. The pair of complex conjugate eigenvalues can be stable as well as
unstable -- unstable again for $\beta_1$ and $\beta_2$ near 1. This is good news:
For typical $\beta_1, \beta_2$ the limit cycle will not turn up in numerical simulations
as it is unstable.
We have not managed to give analytical estimates for stable eigenvalues using CAS so far.

%
%
%

Step 3: Now we show that the limit cycle also exists for $\varepsilon>0$ sufficiently
small. We fix the hyperparameters $\alpha, \beta_1, \beta_2$ and consider the function
$$
F(x,\varepsilon) = \bar T(\bar T(x; \alpha, \beta_1, \beta_2, \varepsilon); \alpha, \beta_1, \beta_2, \varepsilon) - x
$$
Consider a state $\tilde x$ on a 2-cycle for $\tilde \varepsilon=0$ as in step 1, then 
$F(\tilde x, 0)=0$, that is we have a zero of $F$. 
If $\frac {\partial F}{\partial x}(\tilde x, 0)$ is invertible,
then the Implicit Function Theorem  
shows that in a neighbourhood of $\tilde \varepsilon=0$ there exists a
unique zero $x(\varepsilon)$ with $F(x(\varepsilon),\varepsilon)=0$. This zero of $F$
corresponds to a 2-cycle of $\bar T$ with hyperparameter $\varepsilon$.
(We always have $\varepsilon>0$ in Adam, 
but on a non-trivial 2-cycle we have shown
that $v>0$ by the explicit term for $\tilde x_1$ and $\tilde x_2$ in \refb{eq:wtildeExplizit}, 
so even a small reduction of $\varepsilon$ would be allowed.)
Calculating $\det\left(\frac {\partial F}{\partial x}(\tilde x, 0)\right)$ with Maple
we get
\[
4(\beta_1+\beta_2)(\beta_2-1)(\beta_1+1)(2+\beta_1^3\alpha c+(3\beta_2 c\alpha-2\alpha c-2\beta_2)\beta_1^2+3(\beta_2-1)(\alpha c-2\beta_2) \beta_1)
\]
Here $\alpha c$ always appears in combination, with the interpretation that an increase
in the gradient of the objective can be compensated by a decrease in the learning rate.
The leading factors are non-zero because $0<\beta_1,\beta_2<1$, but for any given values 
of $\alpha, \beta_1, \beta_2$ one can zero the final factor with 
\begin{equation}
\label{eq:cException}
 \hat c= \frac {2(\beta_1^2\beta_2+3\beta_1\beta_2^2-3\beta_1\beta_2-1)}{\alpha\beta_1(\beta_1^2+3\beta_1\beta_2-2\beta_1+3\beta_2-3)}
\end{equation}
assuming a non-zero denominator. 
So with the exception of $\hat c(\alpha, \beta_1, \beta_2)$, the limit cycle exists for 
small $\varepsilon$, and by continuous dependence of the eigenvalues on the matrix, 
these limit cycles are also attractive.

Step 4: 
We apply a disturbance estimate to show that $\bar T(x; \alpha, \beta_1, \beta_2, \varepsilon )$ and $ T(t, x; \alpha, \beta_1, \beta_2, \varepsilon)$ have asymptotically the same limit cycles. The following theorem is a variation of \cite[Theorem V.1.]{BockWeiss.2019} and holds for cycles of any integer length.
The proof is very similar and omitted for brevity. The difference between the
variants is that here we do not use an estimate of the type
 $\norm{\Theta(t,\tilde x)} \leq C \beta^t \norm {\tilde x-\xstar}$ where $\xstar$
appears but rather an exponentially decaying term $C \beta^t$. Consequently we cannot
show exponential stability of the 2-limit-cycle but only exponential convergence of
trajectories nearby, with the constant depending on the initial value.

\begin{theorem} 
Let $X\subset \R^n$ be a closed set, $\norm \cdot$ a norm on $\R^n$.
Let  $T : \N_0\times X \to X$ be a mapping which is a contraction w.r.t. the second variable
uniform in $t\in \N_0$, i.e. there exists $L<1$ with
\[
	\norm{T(t,x)-T(t,y)} \leq L \norm{x-y} \quad \forall x,y\in X , t\in\N_0
\]
Furthermore assume that the difference between $T(t+1,\cdot)$ and $T(t,\cdot)$ is 
exponentially bounded: There exist $C\geq 0$, $0<\beta<1$ such that
\[
	\norm{T(t+1,x)-T(t,x)} \leq C \beta^t \quad \forall x\in X , k\in\N_0
\]
Then $T$ has a unique fixed-point $\xstar$ in $X$: $T(t,\xstar) = \xstar$ for all 
$t \in\N_0$. 
For all $x_0 \in X$, the sequence defined by $x_{t+1} = T(t,x_t)$, $t\in\N_0$, 
converges to $\xstar$ exponentially.
\end{theorem}

To apply this theorem to the 2-limit-cycle we have to estimate the difference 
between two iterations of Adam with and without bias correction:
\[
	\norm{T(t+1,T(t,x))- \bar T(\bar T(x))}
\]
We use the fact from \cite{BockWeiss.2019} that 
\[
  \norm {\Theta(t,x)} = 
  \norm {T(t,x)- \bar T(x)} 
  \leq
  C \beta^t \norm {x-\xstar}
\]
as well as 
\[
  \norm {\Theta(t,x)} = 
  \norm {T(t,x)- \bar T(x)} 
  \leq
  C \beta^t \norm {x}
\]
Using this we estimate
\begin{eqnarray*}
  \norm {T(t+1,T(t,x))- \bar T(\bar T(x))}
  	&\leq&
  	\norm {T(t+1,T(t,x))- \bar T(T(t,x))}
  	+
  	\\
  	& & 
  	\norm {\bar T(T(t,x))- \bar T(\bar T(x))}
  	\\
  	&\leq&
  	C \beta^{t+1} \norm{T(t,x)} + L \norm {T(t,x)- \bar T(x)}
\end{eqnarray*}
where $L$ is a local Lipschitz constant near the limit cycle. The Lipschitz continuity exists because $\bar T$ continuously differentiable. We continue the estimate
\begin{eqnarray*}
  	&\leq&
 	C \beta^{t+1} \norm{T(t,x)} + L C \beta^t \norm {x}
 	\\
 	&=&
 	C \beta^{t+1} \norm{\bar T (x) + \Theta(t,x)} + L C \beta^t \norm {x}
 	\\
 	&\leq&
 	C \beta^{t+1} \norm{\bar T (x)} 
 	+ C^2 \beta^{t+1} \beta^{t} \norm{x} + L C \beta^t \norm {x}
 	\\
 	&\leq&
 	\tilde{C} \beta^t \max\{\norm{\bar T (x)}, \norm {x}\}
\end{eqnarray*}
with $\tilde{C} \geq \max\{C^2, LC\}$. As we consider only states $x$ near the limit cycle $\tilde x_1$, $\tilde x_2$
we can locally bound the term $\norm{\bar T (x)}$ by continuity of $\bar T$.

We summarize our findings in the following theorem.
\begin{theorem}
Consider the Adam-Optimizer as defined in Algorithm \ref{alg:ADAM}, 
$f(\weight) = \frac 1 2 c\weight^2$. Then the algorithm
is locally convergent under the assumptions stated \cite{BockWeiss.2019}. 
However if $c\neq \hat c(\alpha, \beta_1, \beta_2)$ with $\hat c(\alpha, \beta_1, \beta_2)$ 
defined in \refb{eq:cException}, there exist solutions that converge to the
2-limit-cycles of the algorithm without bias correction; so 
the algorithm does not converge globally.
\end{theorem}

\section{Numerical Simulations: Discrete Limit cycles}
In \cite[Proposition 3.3]{daSilva.2018} the authors show the existence of a discrete limit cycle for the Adam. This discrete limit cycle depends on the learning rate $\learningrate$ and $\beta_1 = 0$. Therefore we demand $0 < \beta_1 < 1$ this limit cycle does not affect the local convergence proof of \cite{BockWeiss.2019}. But we found in some numerical experiments few limit cycles which alter the convergence of Adam.

First, we will recall the hyperparameter bounding of \cite{BockWeiss.2019}
\begin{align}
\label{eq:Our_inequality}
\frac{\learningrate \max_{i=1}^n (\mu_i)}{\sqrt{\epsilon}} \left( 1-\beta_1 \right) < 2 \beta_1 +2
\end{align}
with $\mu_i$ the $i$-th eigenvalue of the Hessian $\nabla^2f(\weightstar)$. This bounding is marked in the Experiments with a red cross and depicts in every of our Experiments the bifurcation position.

\begin{figure}[!t]
\center
\begin{subfigure}{}
\includegraphics[width=0.5\textwidth]{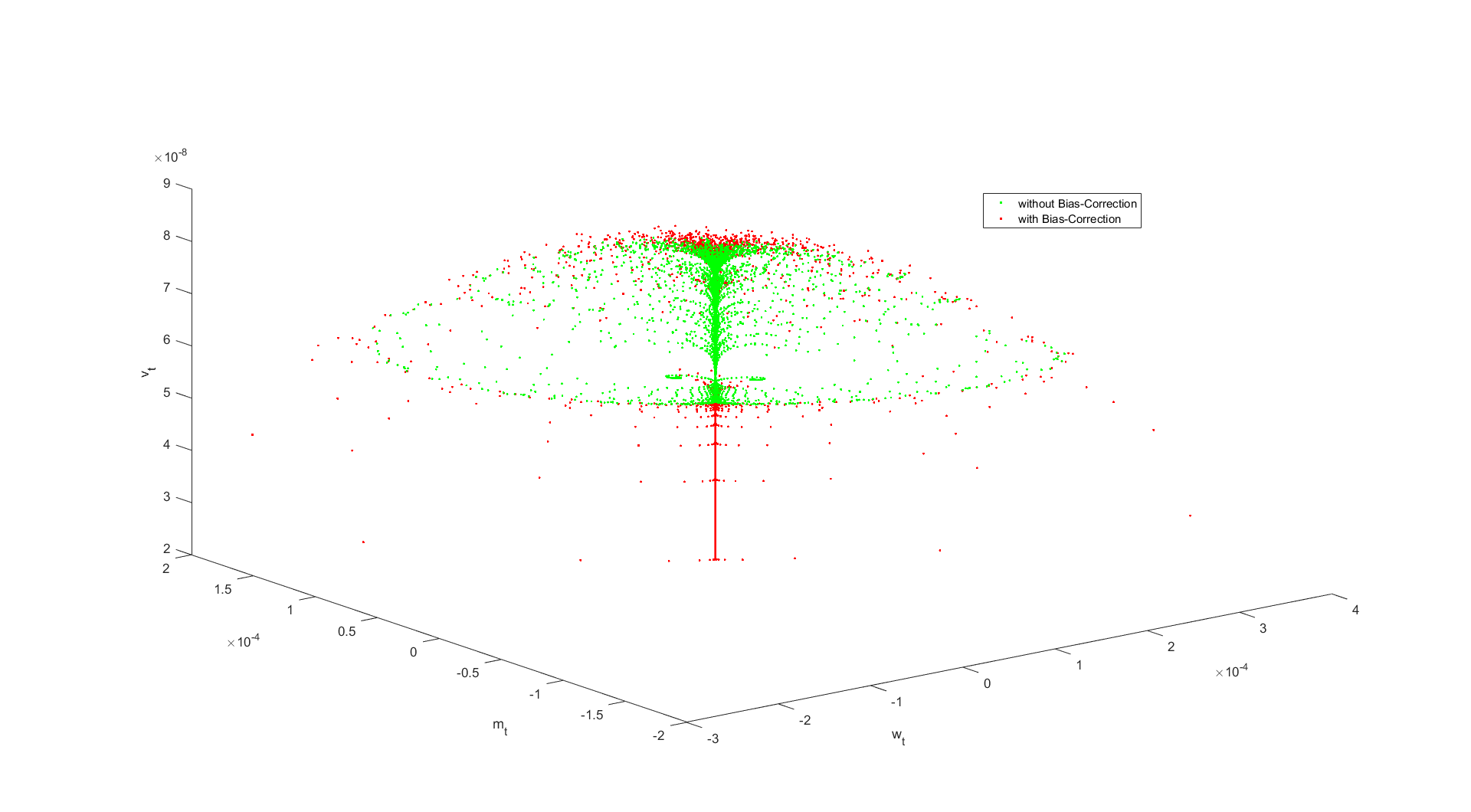}
\end{subfigure}
\begin{subfigure}{}
\includegraphics[width = 0.45\textwidth]{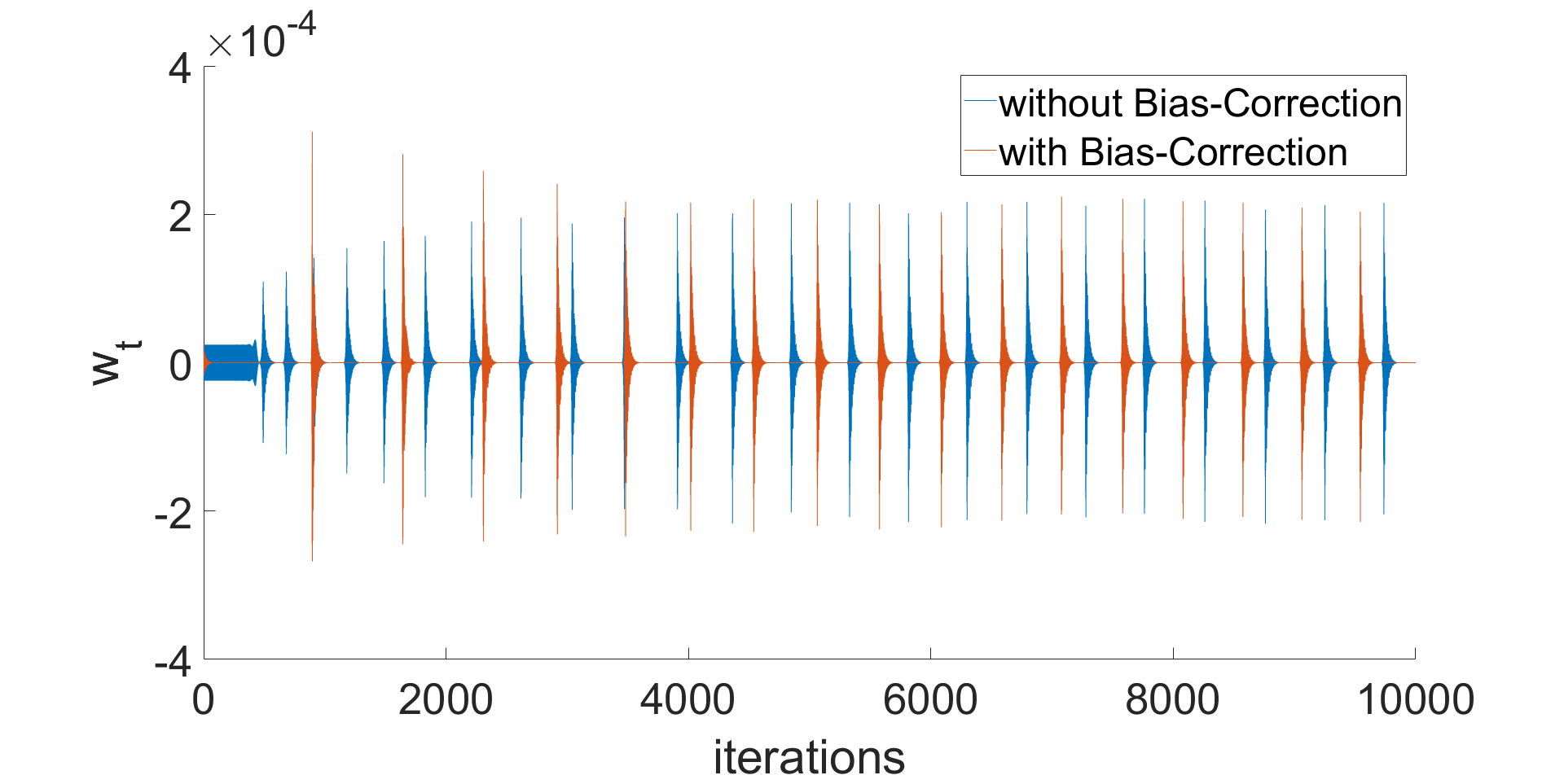}
\end{subfigure}
\caption{Discrete limit cycle with the parameters suggested by \cite{Kingma.2014} (Experiment 1)}
\label{fig:Adam Parameter}
\end{figure}

\begin{figure}[!t]
\center
\includegraphics[width=0.6\textwidth]{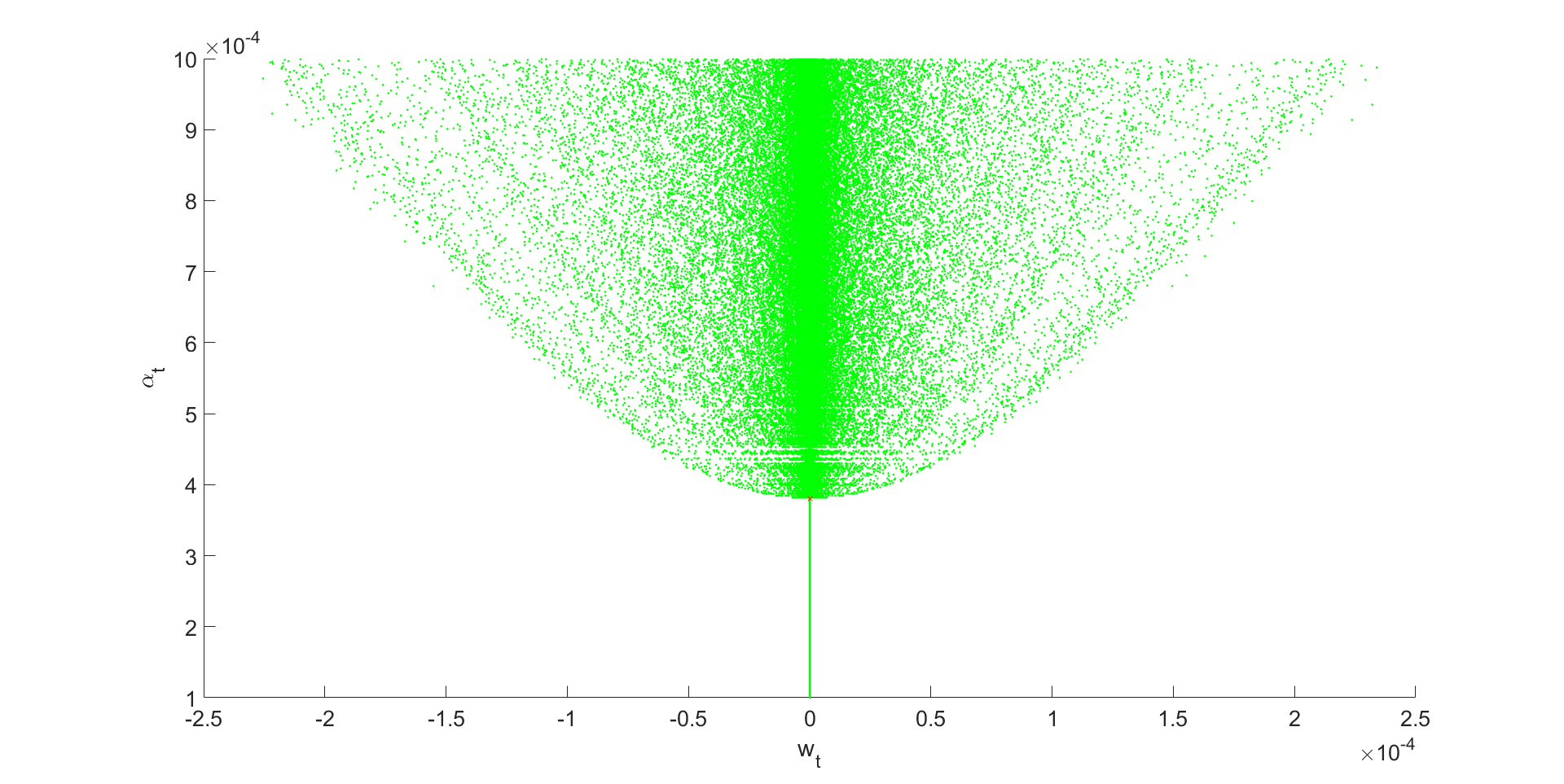} 
\caption{Hopf bifurcation (Experiment 1)}
\label{fig:Adam Bifurcation}
\end{figure}

\begin{table*}
\caption{Parameter of the different limit cycles}
\label{tb:Parameter-Setting}
\centering
\begin{tabular}{c||c|c|c}
 & 1. Experiment & 2. Experiment & 3. Experiment\\
\hline
\hline
$c$ & 10 & 1 & 1\\
\hline
$\learningrate$ & 0.001 & 0.5 & 0.8\\
\hline
$\beta_1$ & 0.9 & 0.2 & 0.5\\
\hline
$\beta_2$ & 0.999 & 0.5 & 0.6\\
\hline
$\varepsilon$ & $10^{-8}$ & $10^{-6}$ & $0.01$\\
\hline
$m_0$ & $-1.281144718*10^{-5}$ & 0 & $0$\\
\hline
$v_0$ & $5.925207756*10^{-8}$ & 0 & $0$\\
\hline
$w_0$ & $2.434174964*10^{-5}$ & $eps$ & $eps$
\end{tabular}
\end{table*}
In the first Experiment, we found a 2-limit-cycle lean on the parameters suggested by \cite{Kingma.2014}. A plot between the three main values $m_t, v_t, w_t$ looks like a fountain and we can see, that the values of $m_t$ and $w_t$ becoming more diffuse by an increasing $v_t$ (see Figure \ref{fig:Adam Parameter} left). By looking closely to $w_t$, it attracts attention that $w_t$ is reaching the solution $0$ but leaving it again (see Figure \ref{fig:Adam Parameter} right). Looking at the eigenvalues of the corresponding Jacobian, it is noticeable that one of them is greater than $1$. Therefore the solution is not a stable 2-limit-cycle. The other real 2-limit-cycles are also not stable. Therefore we reach in Experiment 1 a limit-cycle with a higher order than $2$ (see Figure \ref{fig:Adam Bifurcation}).
In the following we will only consider $\weight$ and $\learningrate$. By the fact that $m_t$ and $v_t$ are only auxiliary variables depending on the history of $\weight$, they are less important than $\weight$ and $\learningrate$. In order to clarify this aspect, reference is made at this point to page $3$ in \cite{Kingma.2014}. There is a definition for $v_t$ without $v_{t-1}$ and thus a definition of $\weight$ without $m$ and $v$ is possible. In addition, the following remark gives an insight into the dependency from $m_t$ to the history of $\weight$ in 2-limit-cycles.
\begin{remark}
We assume a 2-limit-cycle and therefore we can write $m_t = m_{t+2}$. With this knowledge, we can rewrite the $m_t$-update rule.
\begin{align*}
\begin{pmatrix}
-\beta_1 & 1\\
1 & -\beta_1
\end{pmatrix} \begin{pmatrix}
m_t\\
m_{t+1}
\end{pmatrix} = \left( 1-\beta_1\right)\begin{pmatrix}
g(\weight_t)\\
g(\weight_{t+1})
\end{pmatrix}
\end{align*}
Defining $\beta_1 \in (0,1)$ the system is uniquely solvable and thus $m_t$ does not have more information for the system than $\weight$. The same applies to $v_t$.
\end{remark}
If we are iterating over $\learningrate$ from $10^{-4}$ to $0.01$ and declaring $\varepsilon = 10^{-6}$ we reach figure \ref{fig:Adam Bifurcation} and see a Hopf bifurcation. With inequality \refb{eq:Our_inequality} we can calculate exactly the coordinate of the bifurcation (see the red cross in figure \ref{fig:Adam Bifurcation}). 
\begin{align*}
\learningrate_{Bifurcation} = \frac{(2 \beta_1 +2) \sqrt{\epsilon}}{\left( 1-\beta_1\right) \max_{i=1}^n \left(\mu_i\right)} = 0.0038
\end{align*}

In the second Experiment we can see that it is possible that even if we are starting closely to the solution $w_0 = eps$ we are ending in a stable cycle far away.
\begin{figure}[!t]
\center
\begin{subfigure}{}
\includegraphics[width=0.45\textwidth]{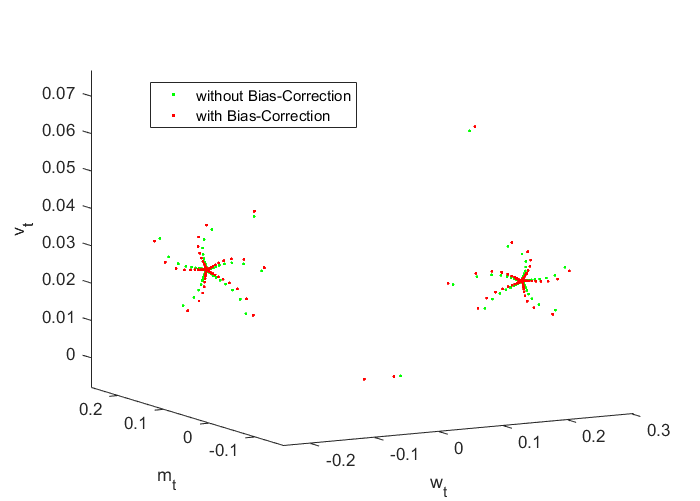}
\end{subfigure}
\begin{subfigure}{}
\includegraphics[trim = 50 270 50 270, clip, width = 0.45\textwidth]{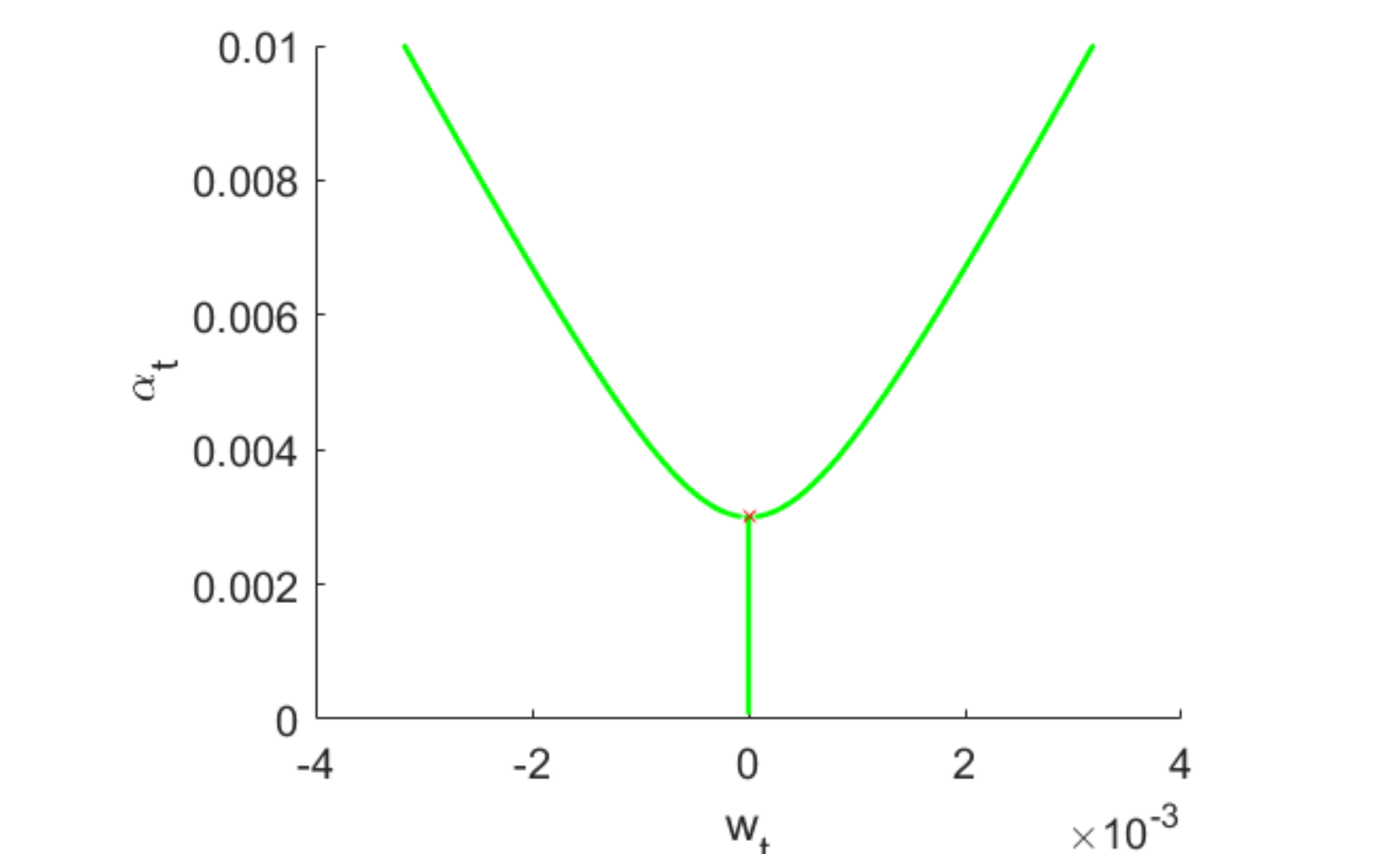}
\end{subfigure}
\caption{Discrete limit cycle starting near the solution (Experiment 2)}
\label{fig:Sternzyklen}
\end{figure}
By iterating over $\learningrate$ from $10^{-4}$ to $0.01$ we can see a pitchfork bifurcation of the Adam. With $\learningrate = 0.5$ stable, the eigenvalues of the corresponding Jacobian are:
\begin{align*}
\lambda_1 &= 0.0113983, \qquad
\lambda_{2,3} = -0.7606667 \pm 0.5465392 \, \text{i}
\end{align*}
In the absolute value all three eigenvalues are lower than $1$ and so we reach a stable 2-limit-cycle between $w_1 = 0.16666$ and $w_2 = -0.16666$.

\begin{figure}[!t]
\begin{subfigure}{}
  \includegraphics[trim = 50 270 50 270, clip, width=0.5\textwidth]{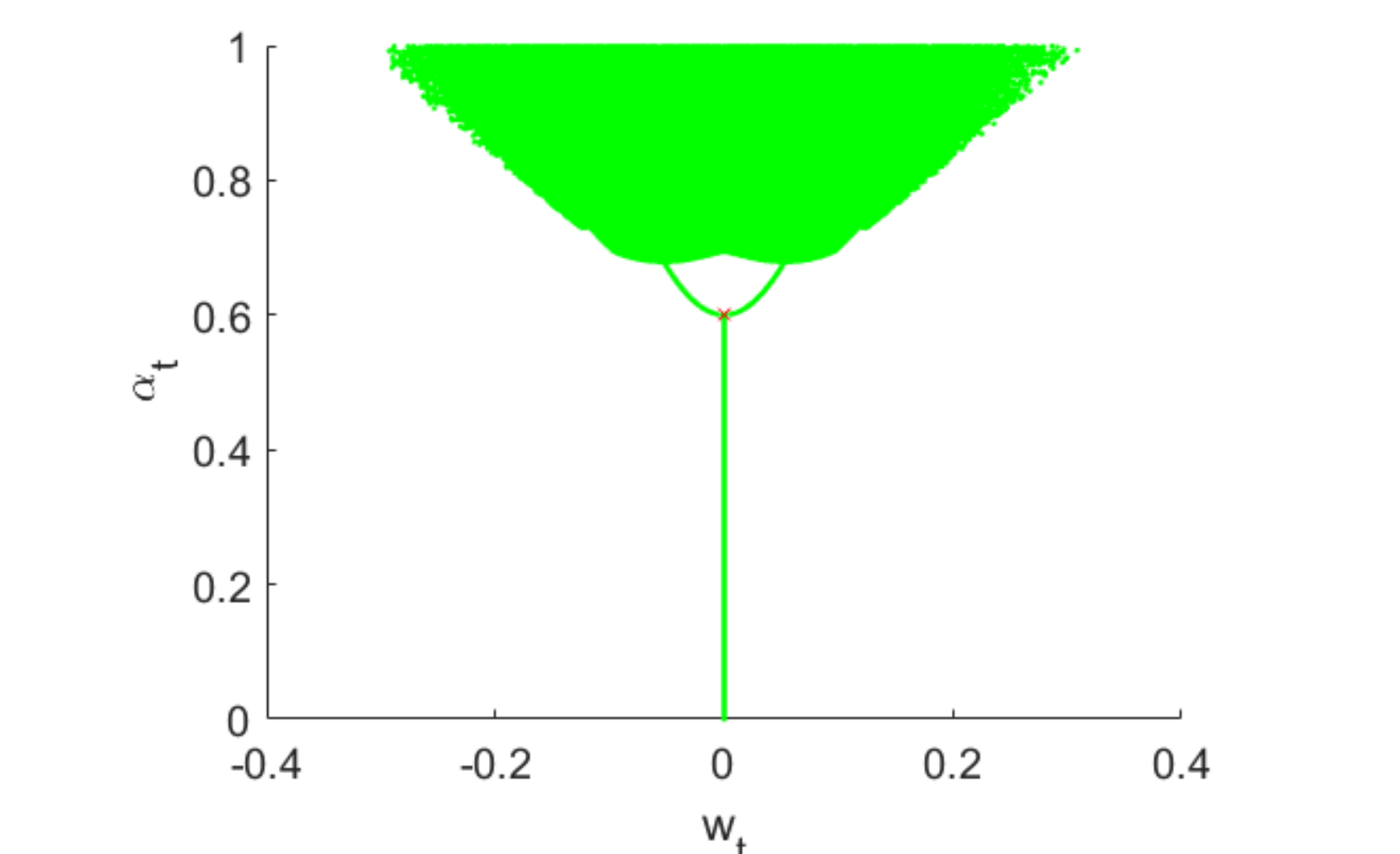}
  \end{subfigure}
  \begin{subfigure}{}
  \includegraphics[width=0.5\textwidth]{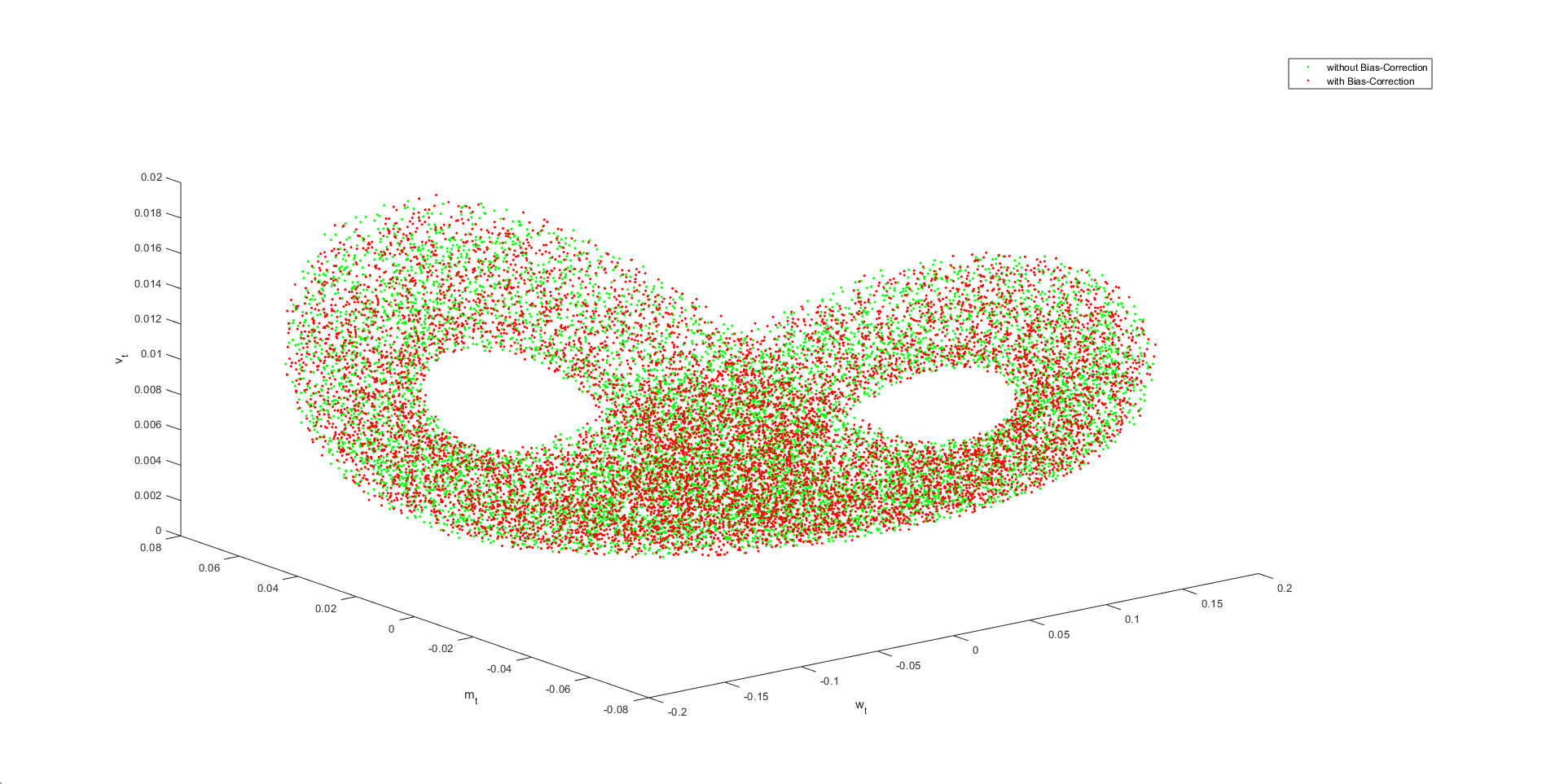}
  \end{subfigure}
\caption{Discrete limit cycle with a large $\varepsilon$ (Experiment 3)}
\label{fig:Experiment_3}
\end{figure}

In contrast to the implicit function argument, Experiment 3 uses a $\varepsilon = 0.01 >> eps$. In Figure \ref{fig:Experiment_3} on the left side one can see that starting at $\learningrate = 0.6$ Adam convergse to a 2-limit-cycle. At around $\learningrate = 0.7$  Adam shows a chaotic behaviour. On the right side one can see the behaviour of the parameters $m,v$ and $w$ at $\learningrate = 0.8$. It visualizes the chaotic behaviour and reminds of the shape of a Lorenz system.

\begin{figure}[!t]
\center
\includegraphics[trim = 50 270 50 270, clip, width=0.6\textwidth]{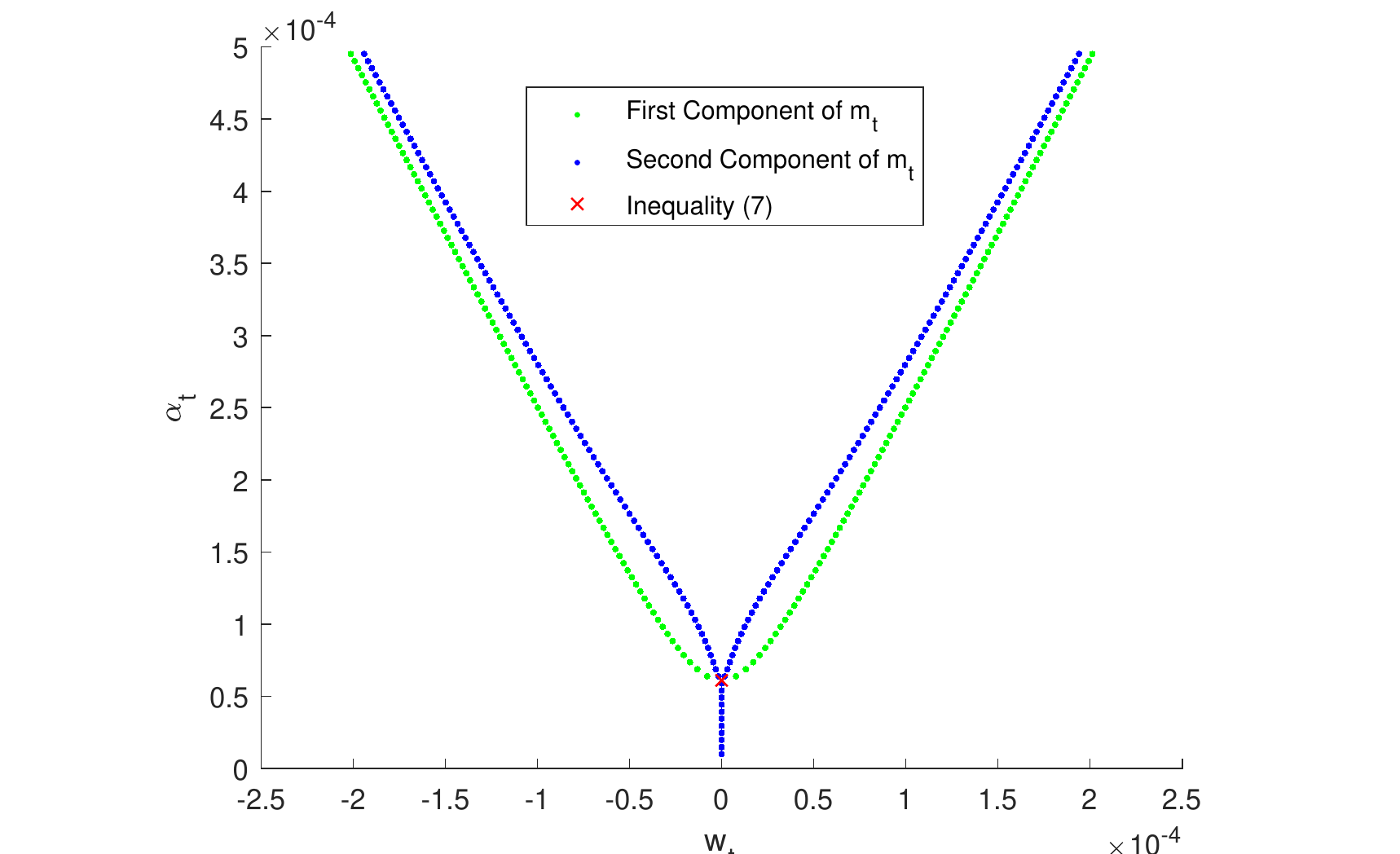}
\caption{Bifurcation in a multidimensional optimization}
\label{fig:Multidimensional Bifurcation}
\end{figure}

Even if we want to minimize a multidimensional problem we can detect such a bifurcation.
For example, with $f(w): \R^2 \rightarrow \R, f(w) = w^\transpose C w$ and 
$$C=\begin{pmatrix}
1.1184 & 0.5841\\
0.5841 & 3.8816
\end{pmatrix} = Q^\transpose \begin{pmatrix}
1 & 0\\
0 & 4
\end{pmatrix} Q$$ 
we obtain Figure \ref{fig:Multidimensional Bifurcation}. 

In every of our experiments is the first bifurcation is exactly on the solved inequality \refb{eq:Our_inequality} (see the red cross in Figure \ref{fig:Adam Bifurcation}, \ref{fig:Sternzyklen}, \ref{fig:Experiment_3} and \ref{fig:Multidimensional Bifurcation}). This result is only an empirical proved assumption and an analytical proof would be desirable.

\section{Conclusion and Discussion}

The results can be extended easily by a diagonalization argument and change of coordinates
to strictly convex quadratic functions 
$\frac 1 2 \weight^\transpose C \weight + b^\transpose \weight + a$ with $\weight\in \R^n$.
But already the scalar quadratic function shows that the Adam dynamics cannot be globally
convergent, even for strictly convex objective function. This implies that there cannot
be a global Lyapunov function. 
Our results seem in contradiction to \cite{De.2018, Chen.2018, Zhou.2018} where
$\varepsilon$-bounded gradients are proven. 
However this contradiction can be explained by the assumption 
$\norm{\nabla f(\weight)}\leq H$ in the cited publications, which we do not use, and the choice of $\beta_1$, $\beta_2$ depending on the bound $\varepsilon$.
So there is still hope for a general convergence result under both assumptions
\[
0 < l I_n \leq \nabla^2 f(\weight) \leq LI_n, 
\qquad
\norm{\nabla f(\weight)}\leq H \qquad \forall \weight\in \R^n
\]  
Furthermore we have investigated only one limit cycle of period 2. More limit cycles
of larger period might exist, so restrictions on objective function and hyperparameters
that eliminate this 2-limit-cycle might miss other or even create other limit cycles.
The computer algebraic methods used in this paper seem hopeless even for period 3, 
so probably completely other methods are necessary.

Our results also have no direct implication on efforts to prove convergence of Adam in 
the incremental mode under additional assumptions. However the lack of a Lyapunov function
suggests that Lyapunov based proofs like \cite{Gadat.2018} for the stochastic heavy ball algorithm, cannot be transferred to Adam in the stochastic or incremental setting.

\printbibliography

\end{document}